\documentclass{article}

\usepackage[nonatbib,final]{nips_2016}

\usepackage[utf8]{inputenc} 
\usepackage[T1]{fontenc}    
\usepackage{hyperref}       
\usepackage{url}            
\usepackage{booktabs}       
\usepackage{amsfonts}       
\usepackage{nicefrac}       
\usepackage{microtype}      

\usepackage{amsmath}
\usepackage{amsfonts}
\usepackage{amssymb}
\usepackage{graphicx}
\usepackage{multirow}
\usepackage{mathtools}

\usepackage{algorithm}
\makeatletter
\newcommand{\removelatexerror}{\let\@latex@error\@gobble}
\makeatother
\usepackage[noend]{algpseudocode}
\usepackage{eqparbox}

\errorcontextlines\maxdimen
\makeatletter
    \newcommand*{\algrule}[1][\algorithmicindent]{\makebox[#1][l]{\hspace*{.5em}\thealgruleextra\vrule height \thealgruleheight depth \thealgruledepth}}%
\newcommand*{\thealgruleextra}{}
\newcommand*{\thealgruleheight}{.75\baselineskip}
\newcommand*{\thealgruledepth}{.25\baselineskip}

\newcount\ALG@printindent@tempcnta
\def\ALG@printindent{%
    \ifnum \theALG@nested>0
        \ifx\ALG@text\ALG@x@notext
        \else
            \unskip
            \addvspace{-1pt}
            \ALG@printindent@tempcnta=1
            \loop
                \algrule[\csname ALG@ind@\the\ALG@printindent@tempcnta\endcsname]%
                \advance \ALG@printindent@tempcnta 1
            \ifnum \ALG@printindent@tempcnta<\numexpr\theALG@nested+1\relax
            \repeat
        \fi
    \fi
    }%
\usepackage{etoolbox}
\patchcmd{\ALG@doentity}{\noindent\hskip\ALG@tlm}{\ALG@printindent}{}{\errmessage{failed to patch}}
\makeatother

\newbox\statebox
\newcommand{\myState}[1]{%
    \setbox\statebox=\vbox{#1}%
    \edef\thealgruleheight{\dimexpr \the\ht\statebox+1pt\relax}%
    \edef\thealgruledepth{\dimexpr \the\dp\statebox+1pt\relax}%
    \ifdim\thealgruleheight<.75\baselineskip
        \def\thealgruleheight{\dimexpr .75\baselineskip+1pt\relax}%
    \fi
    \ifdim\thealgruledepth<.25\baselineskip
        \def\thealgruledepth{\dimexpr .25\baselineskip+1pt\relax}%
    \fi
    \State #1%
    \def\thealgruleheight{\dimexpr .75\baselineskip+1pt\relax}%
    \def\thealgruledepth{\dimexpr .25\baselineskip+1pt\relax}%
}
\newif\ifextendedAbstract
\extendedAbstracttrue

\title{Interpretable Recurrent Neural Networks\\Using Sequential Sparse Recovery}

\author{
  Scott Wisdom$^{1}$, Thomas Powers$^{1}$, James Pitton$^{1,2}$, and Les Atlas$^1$
  \\
  $^1$Department of Electrical Engineering, University of Washington \\
  $^2$Applied Physics Laboratory, University of Washington \\
  \texttt{\{swisdom, tcpowers, pitton, atlas\}@uw.edu} 
}

\begin{document}

\maketitle

\vspace{-15pt}
\begin{abstract}
    Recurrent neural networks (RNNs) are powerful and effective for processing sequential data.
    However, RNNs are usually considered ``black box'' models whose internal structure and learned parameters are not interpretable. In this paper, we propose an interpretable RNN based on the sequential iterative soft-thresholding algorithm (SISTA) for solving the sequential sparse recovery problem, which models a sequence of correlated observations with a sequence of sparse latent vectors. The architecture of the resulting SISTA-RNN is implicitly defined by the computational structure of SISTA, which results in a novel stacked RNN architecture. Furthermore, the weights of the SISTA-RNN are perfectly interpretable as the parameters of a principled statistical model, which in this case include a sparsifying dictionary, iterative step size, and regularization parameters. In addition, on a particular sequential compressive sensing task, the SISTA-RNN trains faster and achieves better performance than conventional state-of-the-art black box RNNs, including long-short term memory (LSTM) RNNs.
\end{abstract}

\ifextendedAbstract
\section{Introduction and relation to prior work}
\vspace{-5pt}


Interpreting the learned features and outputs of machine learning models is problematic. This difficulty is especially significant for deep learning approaches, which are able to learn effective and useful function maps due to their high complexity. Rather than attempt to directly interpret the learned features or outputs of a trained black box deep network, we carefully design the architecture of our deep network based on inference in a probabilistic model. Because the neural network is fully described by probablistic model inference, the network's learned weights and outputs retain their model-based meaning.

Some prior work has taken a similar approach for building intepretable model-based nonrecurrent sparse models. Gregor and LeCun \cite{gregor_learning_2010} proposed the learned iterative soft-thresholding algorithm (LISTA) for sparse coding, which uses learned encoders and decoders to increase the speed and performance of the original ISTA algorithm. Rolfe and LeCun \cite{rolfe_discriminative_2013} created networks from ISTA under a nonnegativity constraint on the 
sparse coefficients.
In this case, the network's nonlinearity
is a rectified linear unit (ReLUs) \cite{nair_rectified_2010}, and network weights are functions of interpretable sparse coding parameters. Kamilov and Mansour \cite{kamilov_learning_2016} learned improved ISTA nonlinearities from data. 
We extend this past work by considering the sequential extension of sparse recovery. 

Regarding past work on human interpretability of RNNs,
Karpathy {et al.} \cite{karpathy2016visualizing} showed that LSTM states produce some meaniningful text annotations. Krakovna and Doshi-Velez \cite{krakovna2016increasing} proposed increasing the interpretability of RNNs by combining hidden Markov models with LSTMs. Unlike this work, our goal is not human interpretability, but model interpretability, which means our proposed SISTA-RNN uses no black box components like LSTMs that are not based on an explicit probabilistic model. Our hope is that our model-based network can provide a better starting point for building human interpretable models.

Our SISTA-RNN also provides a model-based interpretation for an existing black box RNN. A single recurrent layer of the SISTA-RNN is equivalent to another recently proposed architecture, the unitary RNN (uRNN) \cite{arjovsky_unitary_2016,wisdom_full_2016}, except that the SISTA-RNN neither uses a unitary constraint nor complex-valued hidden states. The uRNN has been shown to outperform LSTMs on a variety of tasks. 
This paper is organized as follows. First, we detail our approach to design model-based interpretable deep networks. Then we review conventional stacked RNNs and describe our proposed SISTA-RNN. Finally, we describe our experimental data and summarize our results.

\vspace{-7.5pt}
\section{Building interpretable deep networks}
\vspace{-5pt}

A conventional black box deep network $g$ produces output ${\bf \hat{y}}=g_{\boldsymbol \theta}({\bf x})$ given parameters ${\boldsymbol \theta}$ and input ${\bf x}$.
The parameters ${\boldsymbol \theta}$
are learned on a training set with $I$ input-output pairs 
$\{ ( {\bf x}_{i},{\bf y}_{i} ) \}_{i=1:I}$ 
to minimize a loss function $f$, solving the optimization problem (\ref{eq:opt1})
using stochastic gradient descent.
Usually, the learned parameters ${\boldsymbol \theta}$
are not directly interpretable to a human nor as parameters of a statistical model.
\vspace{-25pt}

\begin{minipage}[t]{0.4\textwidth}
\centering
\begin{align}
& \underset{{\boldsymbol \theta}}{\text{minimize}}
& & \sum_{i=1}^I f\Big( \hat{\bf y}_{i}, {\bf y}_{i}\Big) 
\label{eq:opt1}
\\
& \text{subject to}
& & \hat{\bf y}_{i} =
\;
g_{\boldsymbol \theta}({\bf x}_{i}),
\; i = 1..I.
\nonumber
\end{align}
\end{minipage}
\hspace{0.05\textwidth}
\begin{minipage}[t]{0.5\textwidth}
\centering
\begin{align}
& \underset{\boldsymbol \theta}{\text{minimize}}
& & \sum_{i=1}^I f\Big( \hat{\bf y}_{i}, {\bf y}_{i}\Big) 
\label{eq:opt2}
\\
& \text{subject to}
& & \hat{\bf y}_{i} =
\;
h_{ {\boldsymbol \theta} }({\bf x}_{i}),
\; i = 1..I,
\nonumber
\\
& & & h_{ {\boldsymbol \theta} }({\bf x}_{i})
 = \underset{{\bf z}}{\text{argmin}} \; \mathcal{P}_{ {\boldsymbol \theta}} ({\bf z}, {\bf x}_{i}),
\nonumber
\; i = 1..I.
\end{align}
\end{minipage}

In this paper, we use the idea of {\it deep unfolding} \cite{hershey_deep_2014} to make a modification to the optimization problem (\ref{eq:opt1}), given by (\ref{eq:opt2}).
As in (\ref{eq:opt1}), $f$ is a training loss function as before, but now the network $h_{{\boldsymbol \theta}}$ is a {\it deterministic inference function} parameterized by ${\boldsymbol \theta}$. This inference function $h$ attempts to solve another optimization problem $\mathcal{P}_{ {\boldsymbol \theta} }$ that corresponds to inference in a probabilistic model with parameters ${\boldsymbol \theta}$. Note that parameters ${\boldsymbol \theta}$ can include both model parameters that are part of the probabilistic model and hyper-parameters that are used by the inference function $h$ to optimize $\mathcal{P}$. Because $h_{\boldsymbol \theta}$ attempts to solve an optimization problem that corresponds to a principled probabilistic model, its parameters ${\boldsymbol \theta}$ are perfectly interpretable. For example, we will see that ${\boldsymbol \theta}$ contains objects like a sparsifying signal dictionary and regularization parameters from the problem $\mathcal{P}$.

\vspace{-7.5pt}
\section{Conventional black box stacked RNN}
\vspace{-5pt}

Here we briefly review conventional stacked RNNs. RNNs are often stacked into multiple layers to create more expressive networks 
\cite{
pascanu_how_2013}. The hidden states ${\bf h}_{1:T}$ and output $\hat{\bf y}_{1:T}$ of a stacked RNN are given by
\vspace{-5pt}
\begin{align}
        {\bf h}_t^{(k)}
        =&
        \begin{cases}
            \sigma_{\bf b}\left(
                {\bf W}^{(1)}{\bf h}^{(1)}_{t-1}
                +{\bf V}{\bf x}_t
            \right), 
            & 
            k=1,\\
            \sigma_{\bf b}\left(
                {\bf W}^{(k)}{\bf h}^{(k)}_{t-1}
                +{\bf S}^{(k)}{\bf h}^{(k-1)}_{t}
            \right), 
            & 
            k=2..K,
        \end{cases}
        \label{eq:stackedh}
        \\
        \hat{\bf y}_t
        =&
        {\bf U}{\bf h}_t^{(K)}+{\bf c},
        \label{eq:stackedy}
\end{align}
where $\sigma_{\bf b}$ is a nonlinear activation function, e.g.\ a ReLU function.
The vector ${\bf b}$ denotes optional parameters of the nonlinearity, such as the ReLU threshold.
The parameters of such a stacked RNN are
\begin{equation}
\label{eq:paramsRNN}
{\boldsymbol \theta}_\mathrm{RNN}=
\{{\bf h}_0^{(1:K)},{\bf b}^{(1:K)},{\bf W}^{(1:K)},{\bf V},{\bf S}^{(1:K)},{\bf U},{\bf c}\},
\end{equation}
which are, respectively, initial hidden state, nonlinearity parameters, recurrence matrices, input matrix, cross-layer matrices, output matrix, and output bias.
The left panel of figure \ref{fig:illus} illustrates a stacked RNN and its lattice-like structure.

\begin{figure}[h]
\begin{minipage}{0.49\textwidth}
\includegraphics[width=\linewidth]{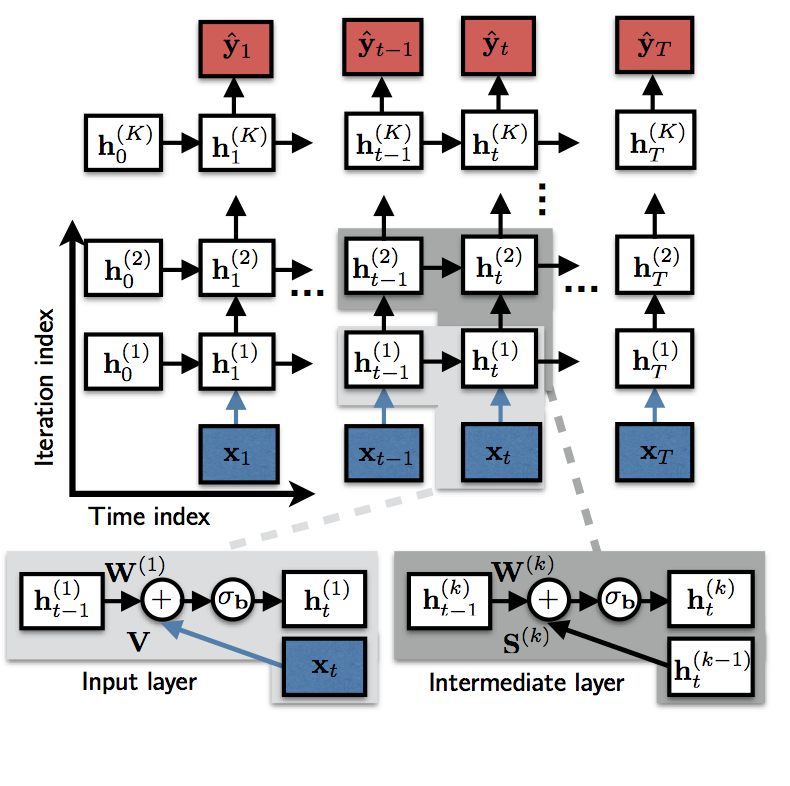}
\end{minipage}
\vline
\begin{minipage}{0.49\textwidth}
\includegraphics[width=\linewidth]{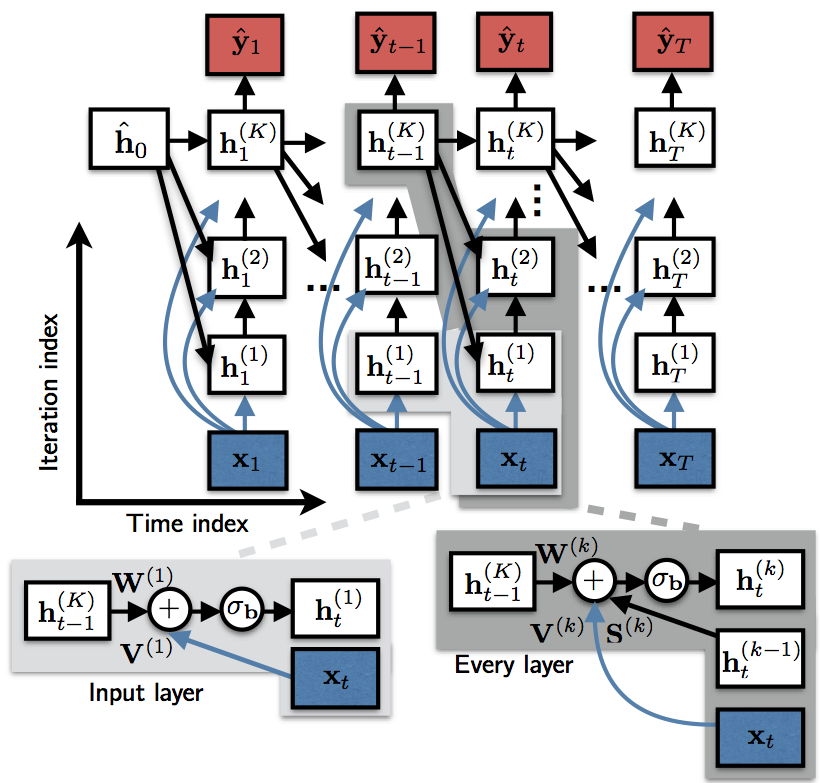}
\end{minipage}
%
\vspace{-5pt}
\caption{Left panel: illustration of conventional black box stacked RNN architecture. Right panel: illustration of interpretable SISTA-RNN architecture derived from the SISTA in algorithm \ref{alg:seqista}.
}
\label{fig:illus}
\vspace{-10pt}
\end{figure}

\section{Interpretable SISTA-RNN}
\vspace{-5pt}

First we describe the specific probablistic model we use.
Then we show that an iterative algorithm for inferring the true denoised signal, the sequential iterative soft-thresholding algorithm (SISTA), corresponds to a particular type of stacked RNN architecture that prescribes different connections between the nodes than a conventional stacked RNN.

The SISTA-RNN uses the following probabilistic model:
\begin{equation}
    \label{eq:modelSISTA}
    \begin{aligned}
        {\bf x}_t
        \sim&\hspace{1pt}
        \mathcal{N}({\bf A}{\bf D}{\bf h}_t,\sigma^2{\bf I}), \\
        p\big({\bf h}_{t}
        \hspace{-2pt}
        \mid
        \hspace{-2pt}
        {\bf h}_{t-1}\big)
        \hspace{-2pt}
        \propto&
        \exp
        \Big\{
            \hspace{-3pt}
            -
            \hspace{-2pt}
            \nu_1\|{\bf h}_t\|_1
            \hspace{-2pt}
            -
            \hspace{-2pt}\frac{\nu_2}{2}\|{\bf D}{\bf h}_t
            \hspace{-2pt}
            -
            \hspace{-2pt}
            {\bf F}{\bf D}{\bf h}_{t-1}\|_2^2
        \Big\}.\hspace{-1pt}.
    \end{aligned}
\end{equation}
That is, every element of an observed sequence is
${\bf x}_t={\bf A}{\bf y}_t+{\bf u}_t$, where ${\bf A}\in\mathbb{R}^{M\times N}$ with $M<N$ is a measurement matrix. ${\bf u}_t$ is Gaussian noise with variance
$\sigma^2$
and the signal ${\bf y}_t$ can be represented using a dictionary 
${\bf D}\in\mathbb{R}^{N\times N}$ 
as 
${\bf y}_t={\bf D}{\bf h}_t$, 
where 
${\bf h}_t$ 
is sparse (because of a Laplace prior with inverse scale $\nu_1$). We also assume that the signal
${\bf y}_t$ is linearly predictable from 
${\bf y}_{t-1}$, such that
${\bf y}_t={\bf F}{\bf y}_{t-1}+{\bf v}_t$,
where ${\bf v}_t$ is zero-mean Gaussian noise with precision $\nu_2$ representing the prediction error.
Minimizing the negative log-likelihood of this model solves the following optimization problem, which corresponds to $\mathcal{P}$ used in equation (\ref{eq:opt2}):
\begin{equation}
\label{eq:seqopt}
    \begin{aligned}
        & \underset{{\bf h}_{1:T}}{\text{minimize}}
        & & \sum_{t=1}^T 
        \Big( 
            \frac{1}{2}
            \big\|{\bf x}_t
            -{\bf A}{\bf D}{\bf h}_{t}
            \big\|_2^2
            +\lambda_1
            \big\|
                {\bf h}_t
            \big\|_1
            +\frac{\lambda_2}{2}
            \|
                {\bf D}{\bf h}_t-{\bf F}{\bf D}{\bf h}_{t-1}
            \|_2^2
        \Big), \\
    \end{aligned}
\end{equation}
with regularization parameters $\lambda_1=2\sigma^2\nu_1$ and $\lambda_2=2\sigma^2\nu_2$.
We dub the iterative algorithm for solving the optimization problem (\ref{eq:seqopt}) the sequential iterative soft-thresholding algorithm (SISTA), 
which
is described in algorithm \ref{alg:seqista} and derived in appendix B. 
The soft-thresholding function $\mathrm{soft}$ is given by
\begin{equation}
\label{eq:soft}
    \mathrm{soft}_{b}(z_n)
    =
    \frac{z_n}{|z_n|}\mathrm{max}(|z_n|-b,0).
    \vspace{-2.5pt}
\end{equation}
The SISTA parameters are
\begin{equation}
\label{eq:paramsSISTA}
{\boldsymbol \theta}_\mathrm{SISTA}=
\{{\bf A},{\bf D},{\bf F},{\bf h}_0,\alpha,\lambda_1,\lambda_2\},
\end{equation}
consisting of measurement matrix, sparsifying dictionary, linear prediction matrix, initial hidden state estimate, SISTA inverse step size, sparse regularization parameter, and regularization parameter for correlation over time. 

\begin{figure}[t]
\removelatexerror
\centering
\begin{minipage}{\linewidth}
\begin{algorithm}[H]
 \caption{Sequential iterative soft-thresholding algorithm (SISTA)\hspace{-5pt}
 }
 \label{alg:seqista}
 \begin{algorithmic}[1]
 \renewcommand{\algorithmicrequire}{\textbf{Input:}}
 \Require
 observed sequence ${\bf x}_{1:T}$, SISTA parameters ${\boldsymbol \theta}_{SISTA}$ from (\ref{eq:paramsSISTA}), and iterations $K$
    \For {$t = 1$ to $T$}
    	\Comment{\# {\it For each time step $t$...}}
        \myState{
        ${\bf h}_t^{(0)}
        \leftarrow
        {\bf D}^{T}{\bf F}{\bf D}\hat{\bf h}_{t-1}$
        }
        \Comment{~~\# {\it Initial estimate for ${\bf h}_t$}}
        \For {$k = 1$ to $K$}
        \Comment{~~\# {\it For $K$ iterations...}}
        \myState{
        %
        ${\bf z}
        \hspace{0pt}
        \leftarrow
        \hspace{0pt}
        \big[
        {\bf I}
        \hspace{-1pt}
        -
        \hspace{-1pt}
        \frac{1}{\alpha}
        {\bf D}^T
        (
        {\bf A}^T{\bf A}
        \hspace{-1pt}
        +
        \hspace{-1pt}
        \lambda_2{\bf I}
        )
        {\bf D}
        \big]
        {\bf h}_t^{(k-1)}
        \hspace{-1pt}
        +
        \hspace{-1pt}
        \frac{1}{\alpha}{\bf D}^T{\bf A}^T
        {\bf x}_t$\hspace{-10pt}
        }
        \label{line:z}
        \Comment{~~~~\# {\it Take a step and enforce}}
        \myState{
        %
        ${\bf h}_t^{(k)}
        \leftarrow
        \mathrm{soft}_{\lambda_1/\alpha}
        \big(
            {\bf z}
            +\frac{\lambda_2}{\alpha}{\bf D}^{T}{\bf F}{\bf D}{\bf \hat{h}}_{t-1}
        \big)$
        }
        \label{line:htk}
        \Comment{~~~~\phantom{\#} {\it sparsity with soft-threshold}}
        \EndFor
    \myState{
    $\hat{\bf h}_t\leftarrow{\bf h}_t^{(K)}$
    }
    \Comment{~\# {\it Assign estimate for ${\bf h}_t$}}
    \label{line:hhat}
    \EndFor
 \Return $\hat{\bf y}_{1:T}={\bf D}\hat{\bf h}_{1:T}$
 \end{algorithmic}
 \end{algorithm}
 \end{minipage}
 \vspace{-10pt}
 \end{figure}

The right panel of figure \ref{fig:illus} illustrates the computational structure of the SISTA algorithm, where the nonlinearity $\sigma_b$ is the soft-thresholding function (\ref{eq:soft}). 
The following equations describe the mapping from SISTA parameters to RNN parameters:

\begin{minipage}{0.6\linewidth}
\begin{align}
            {\bf W}^{(1)}=
            \,&
            \frac{\alpha+\lambda_2}{\alpha}
            {\bf P}
            -\frac{1}{\alpha}
            {\bf D}^T
            ({\bf A}^T{\bf A}+\lambda_2{\bf I}){\bf D}{\bf P},
            \\
            {\bf W}^{(k)}=
            \,&
            \frac{\lambda_2}{\alpha}
            {\bf P}
            \text{,\quad$k>1$,}
            \\
            {\bf S}^{(k)}=\,&
            {\bf I}
            -\frac{1}{\alpha}
            {\bf D}^T
            (
            {\bf A}^T{\bf A}
            +\lambda_2{\bf I}
            )
            {\bf D}
            \text{,\quad$k>1$,}
        \label{eq:equiv2}
\end{align}
\end{minipage}
\begin{minipage}{0.35\linewidth}
\begin{align}
            {\bf V}^{(k)}=\,&\frac{1}{\alpha}{\bf D}^T{\bf A}^T
            \text{,\quad$\forall k$}, 
            \label{eq:equiv1}
            \\
            {\bf P}=\,&{\bf D}^{T}{\bf F}{\bf D} \vphantom{\frac{0}{0}}
            \\
            {\bf U}=\,&{\bf D},\quad {\bf c}={\bf 0} \vphantom{\frac{0}{0}}.
\end{align}
\end{minipage}

Notice the strong similarity to the conventional stacked RNN in the left panel of figure \ref{fig:illus}, with two major differences: the transformed input ${\bf V}{\bf x}_t$ is connected to every layer $k$ (which is similar to residual networks \cite{he2015deep}), and the recurrence connections have a different structure where ${\bf h}_t^{(k)}$ for all layers $k$ takes recurrence from ${\bf h}_t^{(K)}$ in the last layer $K$, instead of from ${\bf h}_t^{(k)}$ in the same layer $k$.

\vspace{-5pt}
\section{Experiment and results}
\vspace{-5pt}

We use a similar experimental setup as Asif and Romberg \cite[\S V.B]{asif_sparse_2014}, which is designed to test sequential compressive sensing algorithms. In this setup, the sequence of signal vectors ${\bf y}_t$ of dimension $N=128$ are the columns of $128\times 128$ grayscale images. Thus, the `time' dimension is actually column index, and all sequences are length $T=128$. The images come from the Caltech-256 dataset \cite{griffin2007caltech}.
We convert the color images to grayscale, clip out centered square regions, and resize to $128\times128$ using bicubic interpolation. The training set consists of $24485$ images, and the validation and test sets consist of $3061$ images each. The columns of each image are observed through a $M\times N$ random measurement matrix ${\bf A}$ with $M=32$ for a compression factor of $4$. The dictionary ${\bf D}$ consists of Daubechies-8 orthogonal wavelets with four levels of decomposition, and we use initial values of $\lambda_1=0.5$ and $\lambda_2=1$ (tuned on the training set), ${\bf h}_0={\bf 0}$, and $\alpha=1$.

We compare our supervised interpretable SISTA-RNN with two conventional supervised black box stacked RNNs: a generic RNN using the soft-thresholding nonlinearity (\ref{eq:soft}) and a stacked LSTM RNN. All stacked RNNs use $K=3$ layers. 
The training loss function $f$ as in (\ref{eq:opt1}) and (\ref{eq:opt2}) is mean-squared error (MSE). The generic RNN and LSTM correspond to $g$ in (\ref{eq:opt1}). All weights in these models are initialized randomly using the suggestion of Glorot and Bengio \cite{glorot2010understanding}. The SISTA-RNN corresponds to $h$ in (\ref{eq:opt2}), which is an inference algorithm that solves the problem $\mathcal{P}$ given by (\ref{eq:seqopt}). The SISTA-RNN is initialized using parameters (\ref{eq:paramsSISTA}) of the unsupervised SISTA algorithm and equations (\ref{eq:equiv1})-(\ref{eq:equiv2}). All SISTA parameters (\ref{eq:paramsSISTA}) are trained.
All code to replicate our results is available at \url{github.com/stwisdom/sista-rnn}.

We also compare to unsupervised baselines, including SISTA with a fixed number of iterations $K=3$ (which is also used as the initialization for the SISTA-RNN). To determine the advantage of allowing more iterations, we also test SISTA run to convergence. Finally, we use $\ell_1$-homotopy \cite{asif_sparse_2014}, an alternative to SISTA. To give $\ell_1$-homotopy an edge against supervised methods, we allow it to use three frames at once for each time step and oracle initialization of the initial hidden state estimate ${\bf h}_0$.

The SISTA-RNN is interpretable. For example, it learns new settings of SISTA parameters as $\lambda_1=3.07$, $\lambda_2=-0.04$ and $\alpha=2.02$. The greater value of $\lambda_1$ suggests increased importance of the sparsity penalty, and greater $\alpha$ suggests a smaller iteration step size. Notice that the learned $\lambda_2$, which is the regularization weight on the $\ell_2$-norm linear prediction error in (\ref{eq:seqopt}), is negative, and thus loses its interpretability as the product of variance $\sigma^2$ and precision $\nu_2$.
This fact suggests a modification to the SISTA-RNN architecture:
add a nonnegativity constraint on $\lambda_2$. Doing so is an example of feeding back information we learn from the trained model to improve its design. 
Figure \ref{fig:images} shows 
example outputs of the different systems and figure \ref{fig:visual} shows visualizations of the initial and learned SISTA parameters. Notice that the dictionary ${\bf D}$ and prediction matrix ${\bf F}$ remain the same, which suggests they are good matches to the data.

Table \ref{tab:results} shows performance results. Notice that the SISTA-RNN achieves the best performance both in terms of MSE and peak signal-to-noise ratio (PSNR), a common objective measure of image quality. Furthermore, notice from the learning curves in figure \ref{fig:learning} that the SISTA-RNN has the additional advantage of training much faster than both the generic RNN and the LSTM. We hypothesize that this improved performance is a consequence of using the principled model-based SISTA initialization. Interestingly, the generic stacked RNN trains faster than the LSTM, which suggests the soft-thresholding nonlinearity is more suitable for this particular task compared to the hyperbolic tangent and sigmoid nonlinearities of the LSTM. 
Also, it is possible that the data may not have long-term dependencies that the LSTM could take advantage of\footnote{Thanks to an anonymous reviewer for this suggestion.}.

\begin{figure}[h]
\vspace{-5pt}
\centering
\begin{minipage}{0.7\textwidth}
\includegraphics[width=\linewidth]{{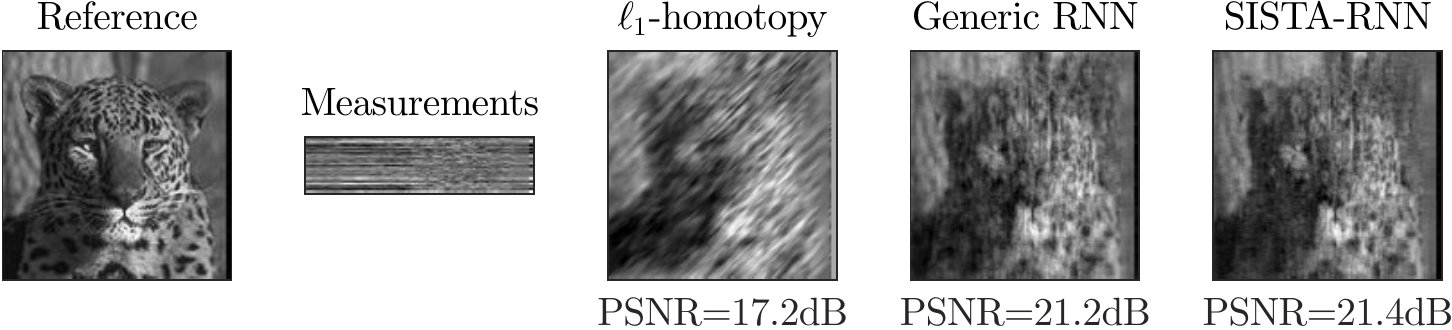}}
\includegraphics[width=\linewidth]{{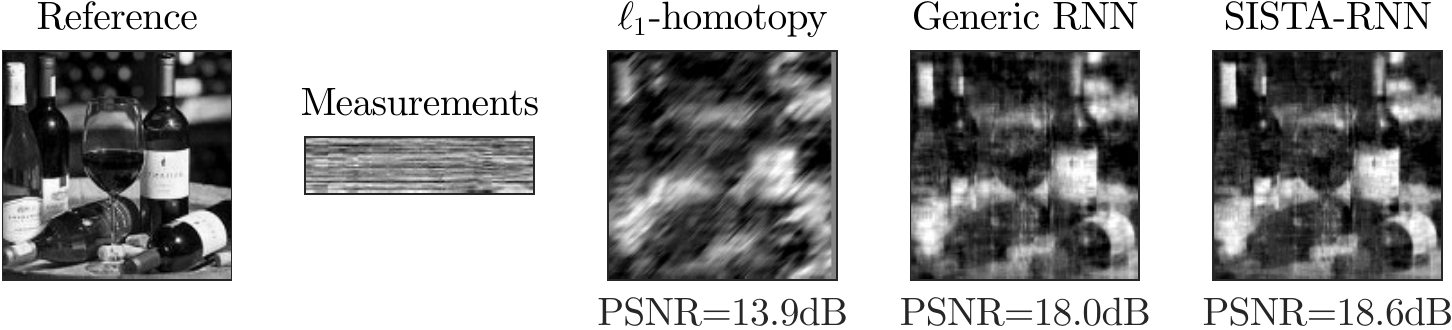}}
\end{minipage}
\vspace{-5pt}
\caption{
Reconstructed images from test set.}
\label{fig:images}
\vspace{-10pt}
\end{figure}

\begin{figure}[h]
    \centering
    \begin{minipage}[t]{0.7\textwidth}
    \includegraphics[trim=0 0 0 0, clip,width=0.49\linewidth]{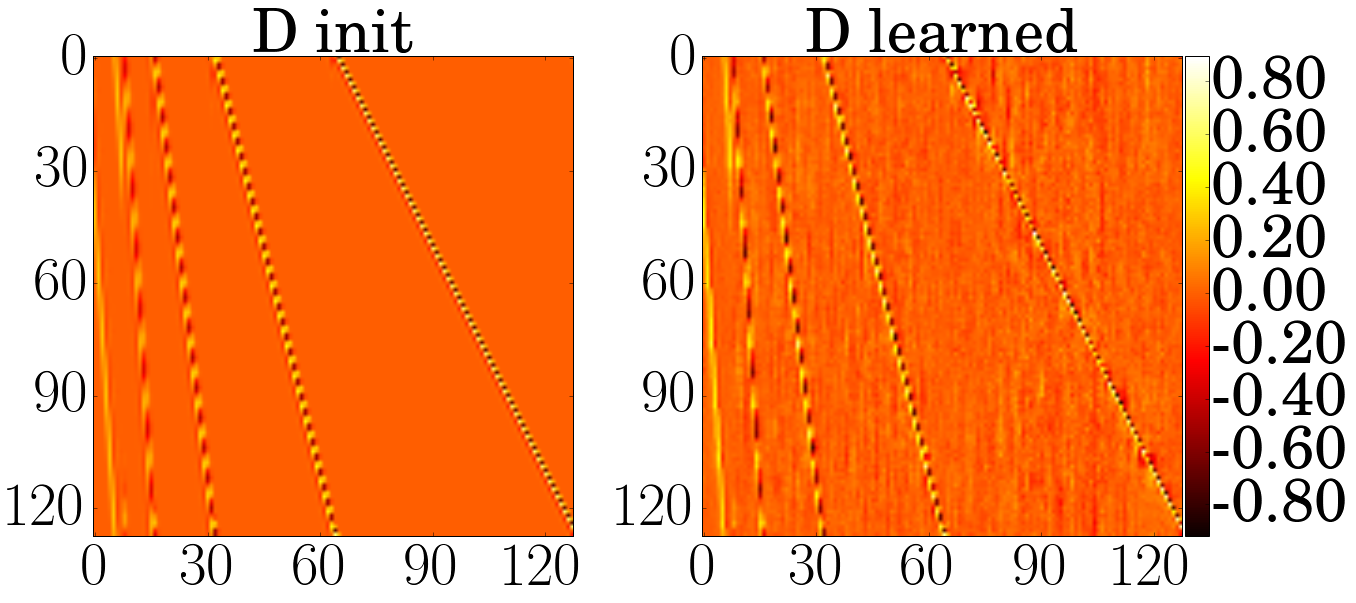}
    \includegraphics[trim=0 0 0 0, clip,width=0.49\linewidth]{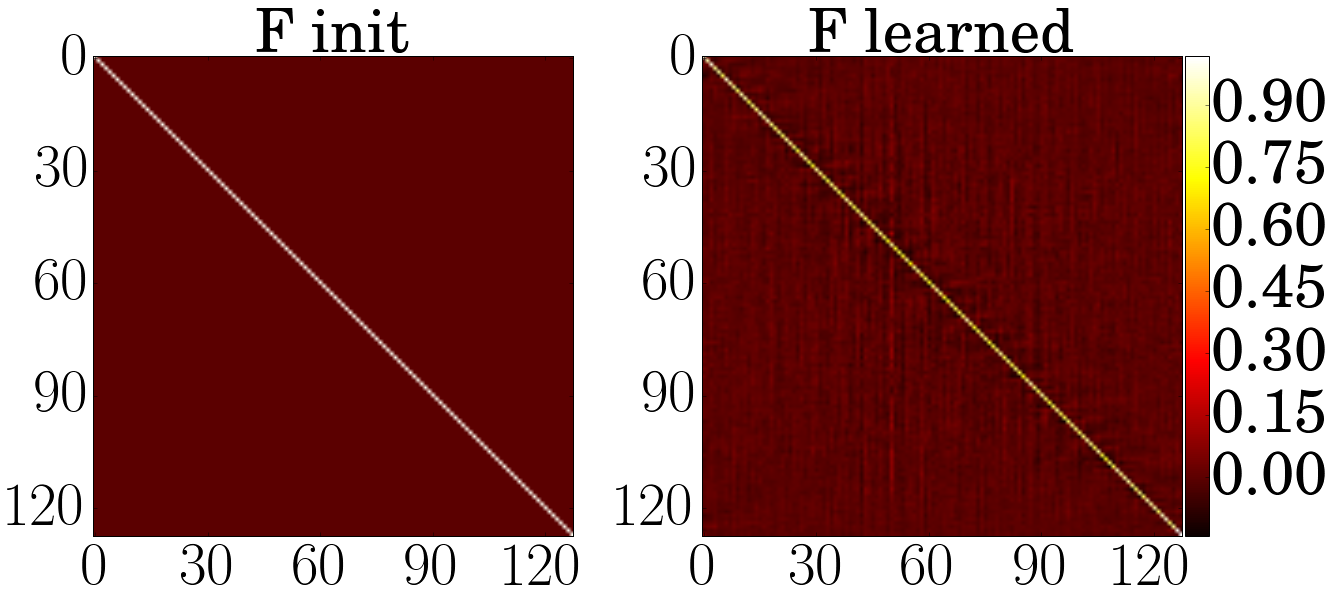} \\
    \includegraphics[trim=0 0 0 0, clip,width=0.49\linewidth]{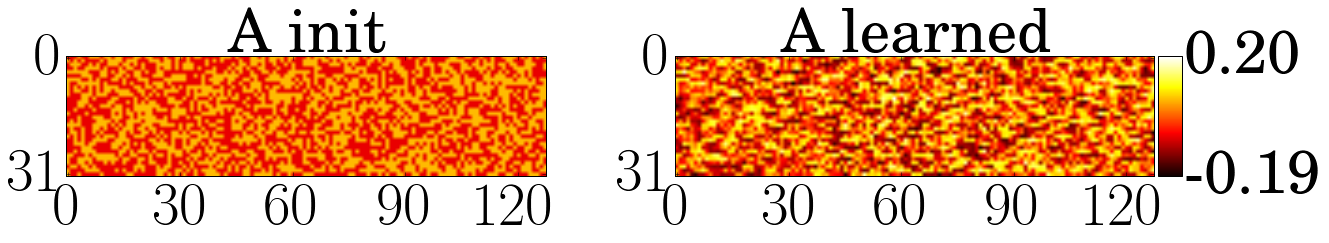}
    \includegraphics[trim=0 0 0 0, clip,width=0.49\linewidth]{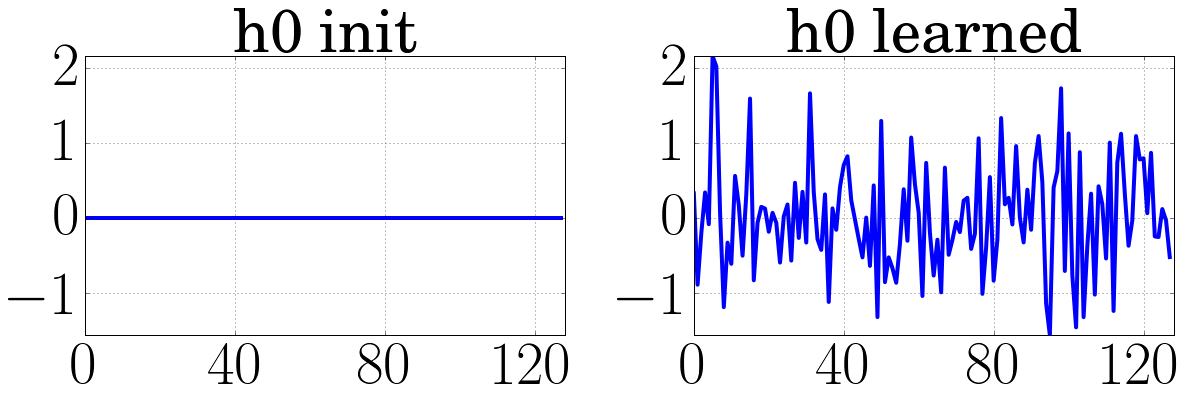}
\end{minipage}
    \caption{Visualizations of some initialized and learned SISTA parameters (\ref{eq:paramsSISTA}) for the SISTA-RNN. See text for settings of $\lambda_1$, $\lambda_2$, and $\alpha$.}
    \label{fig:visual}
    \vspace{-5pt}
\end{figure}

\begin{table}[h]
\centering
\caption{Test set results for sequential sparse recovery in terms of number of iterations $K$, number of training examples $I$, mean-squared error (MSE), and peak signal-to-noise ratio (PSNR).}
\label{tab:results}
\begin{tabular}{clcccc}
         & Algorithm 
         & \# iter.\ $K$ & \# tr.\ $I$ & MSE & PSNR (dB)  \\
         \hline
         \hline
         \multirow{ 5}{5pt}{\rotatebox{90}{Baselines
         }}
         &
         SISTA
         (untrained SISTA-RNN)
         &
         3
         &
         None
         &
         4740 
         & 
         12.1
         \\
         &
         SISTA to convergence
         & 
         $\leq 1825$
         &
         None
         &
         3530 
         & 
         13.4 
         \\
         &
         $\ell_1$-homotopy \cite{asif_sparse_2014}
         (oracle)
         &
         $\leq 314\phantom{0}$
         &
         None
         &
         1490 
         &
         17.1
         \\
         &
         Black box LSTM
         &
         $3$
         &
         $24885$
         &
         727
         &
         20.7
         \\
         &
         Black box generic RNN
         &
         $3$
         &
         $24885$
         &
         720
         &
         20.7
         \\
         \hline
        %
         %
        %
         &
         Proposed: interpretable SISTA-RNN
         &
         $3$
         &
         $24485$
         &
         {\bf 584}
         & 
         {\bf 21.7}
         %
\end{tabular}
\vspace{-15pt}
\end{table}

\begin{figure}[H]
    \centering
    \begin{minipage}{0.75\textwidth}
    \includegraphics[width=\linewidth]{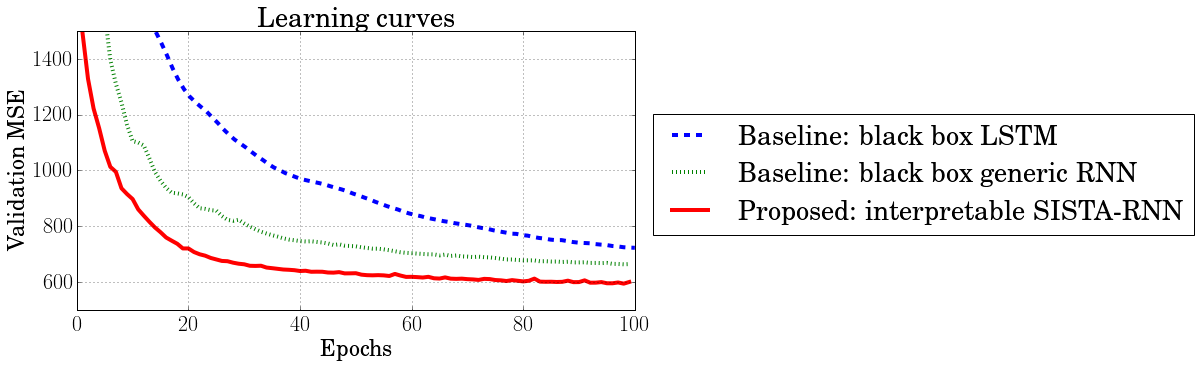}
    \end{minipage}
    \caption{Learning curves for supervised models, showing validation MSE versus training epoch.}
    \label{fig:learning}
    \vspace{-15pt}
\end{figure}


\section{Conclusion}
\vspace{-7.5pt}

We have shown how SISTA, which corresponds to inference in a probabilistic model, can be viewed as a deep recurrent neural network, the SISTA-RNN. The trained weights of the SISTA-RNN maintain their interpretability as parameters of a probablistic model. Furthermore, the SISTA-RNN outperforms two black-box RNN models on a particular image compressive sensing task. Given this promising initial result, we intend to apply the SISTA-RNN to other types of data and further explore how model-based deep networks can assist human interpretability.

\fi



{
\small
\section*{References}


{
\begingroup
\renewcommand{\section}[2]{} 
\bibliographystyle{unsrt_abbrv}
{
\small
\bibliography{nips2016workshop_sista-rnn}
}
\endgroup
}
}

\newpage
\section*{Appendices}
\setcounter{section}{0}
\renewcommand{\thesection}{\Alph{section}}
\section{Sparse recovery}
\label{sec:sparseCoding}

Here we review 
the problem of nonsequential sparse recovery from a single, static observation vector. 
The matrix ${\bf D}\in\mathbb{R}^{N\times N}$ is
a
dictionary whose columns correspond to basis vectors. We make noisy observations of a signal ${\bf s}={\bf D}{\bf h}$ through a measurement matrix ${\bf A}\in\mathbb{F}^{M\times N}$: ${\bf x}={\bf A}{\bf s}+{\boldmath \epsilon}$, where $M<N$ for a compressed sensing problem. 
Sparse recovery solves an optimization problem to find $\hat{\bf h}\in\mathbb{R}^N$ such that the reconstruction ${\bf A}{\bf D}\hat{\bf h}$ is as close as possible to ${\bf x}$ in terms of 
squared error,
subject to a $\ell_1$ penalty:
\begin{equation}
\label{eq:opt}
    \begin{aligned}
        & \underset{\bf h}{\text{min.}}
        & & \frac{1}{2}\|{\bf x}-{\bf A}{\bf D}{\bf h}\|_2^2 + \lambda \|{\bf h}\|_1. \\
    \end{aligned}
\end{equation}
Problem (\ref{eq:opt}) is known as {basis pursuit denoising} (BPDN) \cite{chen_atomic_2001}, which is also equivalent to minimizing the Lagrangian of the {least absolute shrinkage and selection operator} (LASSO) method for sparse recovery \cite{tibshirani_regression_1996}. The $\ell_1$-norm regularization on ${\bf h}$ promotes sparse coefficients, which explain the signal ${\bf s}$ with only a few basis vectors, which are columns of ${\bf D}$.

The LASSO corresponds to a probabilistic model where the observations ${\bf x}$ consist of a deterministic component ${\bf A}{\bf s}={\bf A}{\bf D}{\bf h}$ plus zero-mean Gaussian noise with covariance $\sigma^2{\bf I}$, and each element of ${\bf h}$ has a zero-mean Laplacian prior with scale $\beta$:
\begin{equation}
    \label{eq:modelISTA}
    \begin{aligned}
        {\bf x}
        \sim&
        \mathcal{N}({\bf A}{\bf D}{\bf h},\sigma^2{\bf I}), \\
        h_n
        \sim&
        \mathrm{Laplace}(0,\beta)
        \text{ for $n=1..N$}.
    \end{aligned}
\end{equation}
Minimizing the joint negative log-likelihood of ${\bf x}$ and ${\bf h}$ under this model is equivalent to solving the problem (\ref{eq:opt}) with $\lambda=2\sigma^2/\beta$.

Many algorithms have been proposed, e.g.\ 
\cite{
figueiredo_em_2003,
daubechies_iterative_2004}, for solving the LASSO problem (\ref{eq:opt}).
Here we will focus on ISTA \cite{chambolle_nonlinear_1998,daubechies_iterative_2004}, which is a proximal gradient method that consists of $K$ iterations of soft-thresholding. The basic ISTA algorithm is described in algorithm \ref{alg:ista}, where $1/\alpha$ is a step size and $\mathrm{soft}_b({\bf z})$ of a vector ${\bf z}$ denotes application of the soft-thresholding operation (\ref{eq:soft}) with real-valued threshold $b$ to each element $z_n$ of ${\bf z}$.
\begin{figure}[h]
\removelatexerror
\begin{algorithm}[H]
 \caption{Basic iterative soft-thresholding algorithm (ISTA)}
 \label{alg:ista}
 \begin{algorithmic}[1]
 \renewcommand{\algorithmicrequire}{\textbf{Input:}}
 \Require
 observations ${\bf x}$, measurement matrix ${\bf A}$, dictionary ${\bf D}$, initial coefficients ${\bf h}^{(0)}$
  \For{$k = 1$ to $K$}
    \State
    $
    \makebox[0pt][l]{{\bf z}}\phantom{{\bf h}^{(k)}}
    \leftarrow
    ({\bf I}-\frac{1}{\alpha}{\bf D}^T{\bf A}^T{\bf A}{\bf D}){\bf h}^{(k-1)}
    +
    \frac{1}{\alpha}{\bf D}^T{\bf A}^T
    {\bf x}
    $
    \State
    $
    {\bf h}^{(k)}
    \leftarrow
    \mathrm{soft}_{\lambda/\alpha}
    \left(
    {\bf z}
    \right)
    $
  \EndFor \\
 \Return ${\bf h}^{(K)}$
 \end{algorithmic}
 \end{algorithm}
 \vspace{-10pt}
 \end{figure}
 
More efficient variants of ISTA have been proposed that use an adaptive step size $\alpha^{(k)}$ and iteration-dependent $\lambda^{(k)}$, such as fast ISTA (FISTA) \cite{beck_fast_2009}, which combines Nesterov's adaptive step size method \cite{nesterov_introductory_2013} with a gradual decrease in $\lambda^{(k)}$, and sparse reconstruction by separable approximation (SparSA) \cite{wright_sparse_2009}, which combines Barzilai-Borwein gradient step size adjustment \cite{barzilai_two-point_1988} with a gradual decrease in $\lambda^{(k)}$.

\section{Derivations of ISTA and SISTA for sparse recovery}
\label{sec:ista}

In this section we derive the ISTA algorithms for nonsequential and sequential sparse recovery.

\subsection{ISTA for nonsequential sparse recovery}
Here we review the derivation of an iterative shrinkage and thresholding algorithm (ISTA) for any optimization problem of the form 
\begin{equation}
    \begin{aligned}
        & \underset{\bf h}{\text{min.}}
        & & L({\bf h}) + \lambda R({\bf h}), \\
    \end{aligned}
    \label{eq:optgen}
\end{equation}
of which (\ref{eq:opt}) is an example with $L({\bf h})=\frac{1}{2}\|{\bf x}-{\bf A}{\bf D}{\bf h}\|_2^2$ 
and $R({\bf h})=\|{\bf h}\|_1$. Our presentation follows Wright et al. \cite{wright_sparse_2009}.

Often problems of the form (\ref{eq:optgen}) can be difficult to solve, such as the $\ell_1$-regularized $\ell_2$ problems we consider in this letter. ISTA provides a solution to problem (\ref{eq:optgen}) using an iterative sequence of subproblems, each of which can be solved efficiently:
\begin{equation}
    \begin{aligned}
        & 
        {\bf h}^{(k)}=\underset{\bf h}{\text{argmin}}
        & & 
        ({\bf h}-{\bf h}^{(k-1)})^T {\bf G}({\bf h}^{(k-1)}) 
        +\frac{\alpha}{2}\|{\bf h}-{\bf h}^{(k-1)}\|_2^2 
        + \lambda R({\bf h}),
    \end{aligned}
    \label{eq:ista1}
\end{equation}
where ${\bf G}({\bf h}^{(k-1)})=\nabla L({\bf h}^{(k-1)})$ is the gradient of the loss function at the previous estimate ${\bf h}^{(k-1)}$. An equivalent form of (\ref{eq:ista1}) is
\begin{equation}
    \begin{aligned}
        & 
        {\bf h}^{(k)}=\underset{\bf h}{\text{argmin}}
        & & 
        \frac{1}{2}\|{\bf h}-{\bf u}^{(k-1)}\|_2^2
        + \frac{\lambda}{\alpha} R({\bf h}),
    \end{aligned}
    \label{eq:ista2}
\end{equation}
where
\begin{equation}
    {\bf u}^{(k-1)}
    =
    {\bf h}^{(k-1)}
    -
    \frac{1}{\alpha}{\bf G}({\bf h}^{(k-1)}).
    \label{eq:ista3}
\end{equation}

For the specific choice of $L({\bf h})=\frac{1}{2}\|{\bf x}-{\bf A}{\bf D}{\bf h}\|^2_2$, the gradient is
\begin{equation}
    {\bf G}({\bf h})
    =
    -{\bf D}^T{\bf A}^T({\bf x}-{\bf A}{\bf D}{\bf h}).
    \label{eq:ista4}
\end{equation}
When $R({\bf h})=\|{\bf h}\|_1=\sum_{n=1}^N |h_n|$, it is not differentiable at any $h_n=0$, and thus has subdifferential
\begin{equation}
    \frac{\delta}{\delta h_n}R({\bf h})
    =
    \begin{cases}   
        -1,     & h_n < 0\\
        [-1,1], & h_n = 0\\
        1,      & h_n > 0,
    \end{cases}
    \label{eq:ista5}
\end{equation}
where $[a,b]$ denotes a continuous range from $a$ to $b$.
Combining (\ref{eq:ista2}-\ref{eq:ista5}), taking the gradient with respect to ${\bf h}^{(k)}$, setting this gradient equal to $0$, and solving for ${\bf h}^{(k)}$ yields
\begin{equation}
    {\bf h}^{(k)}
    =
    \mathrm{soft}_{\lambda/\alpha}
    \left(
        ({\bf I}-\frac{1}{\alpha}{\bf D}^T{\bf A}^T{\bf A}{\bf D}){\bf h}^{(k-1)}
        +
        \frac{1}{\alpha}{\bf D}^T{\bf A}^T
        {\bf x}
    \right),
    \label{eq:ista_update}
\end{equation}
which is the ISTA update step in algorithm \ref{alg:ista}.

\subsection{ISTA for sequential sparse recovery}


Here we derive ISTA to solve the optimization problem (\ref{eq:seqopt}) for sequential sparse recovery. Using the variables
\begin{align}
    {\bf \bar{D}}
    =&
    \left[
    \begin{array}{c}
         {\bf A}{\bf D}  \\
         -\sqrt{\lambda_2}{\bf D}
    \end{array}
    \right],
    \quad
    {\bf \bar{x}}_t
    =
    \left[
    \begin{array}{c}
         {\bf x}_t  \\
         -\sqrt{\lambda_2}{\bf F}{\bf D}{\bf h}_{t-1}
    \end{array}
    \right],
    \label{eq:replace}
\end{align}
we can write the problem (\ref{eq:seqopt}) in an equivalent form
\begin{equation}
    \begin{aligned}
        & \underset{{\bf h}_{1:T}}{\text{min.}}
        & & \sum_{t=1}^T 
        \Big( 
            \frac{1}{2}
            \big\|{\bf \bar{x}}_t
            -{\bf \bar{D}}{\bf h}_{t}
            \big\|_2^2
            +\lambda_1
            \big\|
                {\bf h}_t
            \big\|_1
        \Big).
    \end{aligned}
    \label{eq:seqopt_equiv}
\end{equation}
Problem (\ref{eq:seqopt_equiv}) is of similar form to problem (\ref{eq:opt}) for each time step. Thus, for each time step $t$, given that we have computed an estimate $\hat{\bf h}_{t-1}$ for the previous time step, we can plug in the quantities ${\bf \bar{D}}$ and ${\bf \bar{x}}_t$ from (\ref{eq:replace}) for ${\bf AD}$ and ${\bf x}$ within the ISTA update (\ref{eq:ista_update}). If we use the estimate after the $K$th iteration as the estimate of time step $t-1$, $\hat{\bf h}_{t-1}={\bf h}_{t-1}^{(K)}$, and if we initialize the optimization for time step $t$ using the prediction from the previous time step's estimate, ${\bf h}_t^{(0)}={\bf D}^T{\bf F}{\bf D}\hat{\bf h}_{t-1}$, then iterative optimization is equivalent to SISTA in algorithm \ref{alg:seqista}.

\end{document}